\documentclass[runningheads]{llncs}
\usepackage{times}  
\usepackage{helvet}  
\usepackage{courier}  
\usepackage{algorithm}
\usepackage{algorithmic}
\usepackage{amssymb}
\usepackage{tikz}
\usetikzlibrary{positioning}
\usetikzlibrary{arrows}
\usetikzlibrary{shapes}
\usepackage{graphicx}
\usepackage{amsmath}
\usepackage{multicol}
\usepackage{multirow}
\usepackage{subcaption}
\usepackage{makecell}
\usepackage{hyperref}
\usepackage{newfloat}
\usepackage{listings}
\usepackage[T1]{fontenc}
\usepackage{color}
\usepackage{graphicx}
\begin{document}
\title{Differentiable Inductive Logic Programming in High-Dimensional Space}
%
%
\author{Stanisław J. Purgał \inst{2}\orcidID{0009-0000-1198-9430} \and
David M. Cerna\inst{1}\orcidID{0000-0002-6352-603X} \and
Cezary Kaliszyk\inst{3}\orcidID{0000-0002-8273-6059}}
\authorrunning{S.J. Purgał et al.}
%
\institute{Czech Academy of Sciences Institute of Computer Science, Prague, Czechia 
\email{dcerna@cs.cas.cz}\\
 \and
University of Innsbruck, Innsbruck, Austria\\
\email{sjpurgal@gmail.com}
\and 
University of Melbourne, Melbourne, Australia\\
\email{cezary.kaliszyk@unimelb.edu.au}}

\maketitle              
\begin{abstract}
 Synthesizing large logic programs through symbolic Inductive Logic Programming
(ILP) typically requires intermediate definitions. However, cluttering the hypothesis space with invented predicates typically degrades performance. In contrast,  gradient descent provides an
efficient method to find solutions within high-dimensional spaces; a property not  fully exploited
by neuro-symbolic ILP approaches. We propose extending the \textit{differentiable} ILP framework by large-scale (extending its small-scale) predicate invention to emulate search through a high-dimensional space, and thus allowing us to exploit the efficacy of gradient descent. We show that large-scale predicate invention is beneficial to differentiable inductive synthesis and results in learning  capabilities beyond existing neuro-symbolic ILP systems. Furthermore, we achieve these results without specifying the precise structure of the solution within the \textit{inductive bias}.
\end{abstract}

\section{Introduction}

Neuro-symbolic ILP is quickly becoming one of the most important research domains in inductive synthesis \cite{CropperD22}. Such systems can aid explainability research by providing logical representations of what was learned and providing noise-handling capabilities to symbolic learners. Systems such as $\delta$ILP can consistently learn solutions for many standard inductive synthesis problems \cite{dilp1,SPD2023}. Nonetheless, searching through the hypothesis space remains a challenging task. To deal with this difficulty, inductive learners introduce problem-specific restrictions that reduce the size of the respective search space \cite{ShindoNY21,SenCRG22}; what is commonly referred to as  \textit{language bias}. While this is conducive to solving simple learning tasks, more complex synthesis tasks or tasks where the required language bias is not easily specifiable remain a formidable challenge.

Predicate invention (PI) is a technique that enables the creation of new predicates, thereby adding shortened programs to the hypothesis space and circumventing the limitations imposed by the user-provided background knowledge. However, in purely symbolic inductive synthesis, reliance on large-scale PI is avoided, as it is time- and space-wise too demanding \cite{MuggletonRPBFIS12}, and has only been used effectively in very restricted settings~\cite{oneshotlearning2023}. 

This paper introduces a novel approach to neuro-symbolic inductive synthesis, compatible with large-scale predicate invention, that leverages the power of the highly influential differentiable ILP \cite{dilp1}. We show that a few changes to the original architecture is enough to train models using a reduced language bias that tests with a significantly higher success rate. Our extension synthesizes a user-provided number of invented predicates during the learning process. Large-scale PI is either intractable for most systems due to memory requirements or results in a performance drop as it clutters the hypothesis space. 
In contrast, gradient descent methods generally benefit from large search spaces (high dimensionality). Thus, we propose introducing a large number of invented predicates to improve performance.
We evaluate the approach on several standard ILP tasks (many derived from~\cite{dilp1}), including several which existing neuro-symbolic ILP systems find to be a significant challenge (see \textit{Hypothesis 1}). 

Solutions found by our extension of $\delta$ILP, in contrast to the usual ILP solutions, include large numbers of invented predicates. We posit the usefulness of large-scale PI for synthesizing complex logic programs. While adding many invented predicates can be seen as a duplication of the search space and, therefore, equivalent to multiple initializations of existing neuro-symbolic ILP systems, we demonstrate that our extension of $\delta$ILP easily outperforms the re-initialization approach on a particularly challenging task (see \textit{Hypothesis 2}). We compare to the most relevant existing approach, $\delta$ILP (presented in~\cite{dilp1}). 

Unlike the experiments presented in~\cite{dilp1}, which specify the solution's precise structure, we assume a generic shape for all predicate definitions. Thus, our experiments force the learner to find both the correct predicates to reference within an invented predicate's definition and the structure of the definition. In this experimental setting, our approach is on par with $\delta$ILP and, for challenging tasks, outperforms it. In particular, we outperform $\delta$ILP on tasks deemed difficult in~\cite{dilp1} such as \textit{0=X mod 3} and \textit{0=X mod 5}. 

Furthermore, we propose an adjusted measure of task difficulty. In~\cite{dilp1}, the authors proposed the number of \textit{learned} predicate definitions as a measure of learning complexity. While our results do not contradict this assertion, it is more precise to focus on the relation between input variables and body-only variables, i.e., we solve \textit{0=X mod 5} (requires learning four predicate definitions) consistently but perform poorly on the seemingly simpler task  $Y=X+4$ which requires learning two predicates (\textit{Hypothesis 3}). Improving our understanding of task difficulty will aid future investigations.

Our contributions are as follows: \textbf{(i)} an extension of $\delta$ILP capable of large-scale PI (\textit{Hypothesis 1})\footnote{Our implementation can be found at the following repository: \href{https://github.com/Ermine516/DILP2}{\underline{{\color{blue}github.com/Ermine516/DILP2}}}.}, \textbf{(ii)} experimental verification of improved performance on challenging tasks (\textit{Hypothesis 1}), \textbf{(iii)} experimental verification that large-scale PI differs from weight re-initialization (\textit{Hypothesis 2}), \textbf{(iv)} proposing a novel complexity criterion and experimentally validating it (\textit{Hypothesis 3}).

\section{Related Work}
\label{sec:relatedwork}
We briefly introduce Inductive logic programming~\cite{CropperD22}, cover aspects of $\delta$ILP~\cite{dilp1} directly relevant to our increase in dimensionality, and compare our approach to related systems inspired by $\delta$ILP. We assume familiarity with basic logic and logic programming; see \cite{deraedt2008}. 

\subsection{Inductive Logic Programming (ILP)}
ILP is traditionally a form of symbolic machine learning whose goal is to derive explanatory hypotheses from sets of examples (denoted $E^+$ and $E^-$) together with background knowledge (denoted $\mathit{BK}$). Investigations often represent explanatory hypotheses as logic programs~\cite{CropperM21,metagol,ILASP,HopperSDC2021,FOIL1990}. A benefit of this approach is that only a few examples are typically needed to learn an explanatory hypothesis~\cite{DaiMWTZ17}. 

The most common learning paradigm implemented within ILP systems is \textit{learning from entailment}~\cite{deraedt2008}. The systems referenced above, including  $\delta$ILP, use this paradigm, which is succinctly stated as follows: A hypothesis $H$ explains  $E^+$ and $E^-$ through the $\mathit{BK}$, if
$$ \forall e\in E^+, \mathit{BK}\wedge H\models e\hspace{1em} \textrm{and} \hspace{1em} \forall e\in E^-, \mathit{BK}\wedge H\not \models e$$

Essentially, the hypothesis, together with the background knowledge, entails all the positive examples and none of the negative examples. In addition to the learning paradigm, one must consider how to search through the \textit{hypothesis space}, the set of logic programs constructible using definitions from the \textit{BK} together with the predicates provided as examples. Many approaches exploit \textit{subsumption} ($\leq_{sub}$), which has the following property in relation to entailment: $H_1\leq_{sub} H_2 \Rightarrow H_1\models H_2$ where $H_1$ and $H_2$ are plausible hypotheses. Subsumption provides a measure of specificity between hypotheses and, thus, is used to measure progress. The FOIL~\cite{FOIL1990} approach (\textit{top-down}) iteratively builds logic programs using this principle. \textit{Bottom-up} approaches, i.e., Progol~\cite{Muggleton95}, build the subsumptively most specific clause for each positive example and use FOIL to extend more general clauses towards it.

The ILP system \textit{Metagol}~\cite{metagol}  implements the meta-learning approach to search. It uses second-order Horn templates to restrict and search the hypothesis space. An example template would be $P(x,y)\mbox{:-} Q(x,z), R(z,y)$ where $P, Q,$ and $R$ are variables ranging over predicate symbols. This approach motivates the \textit{template} representation of $\delta$ILP and our work.
\subsection{Differentiable ILP}
\label{subsec:dilp}
Given that $\delta$ILP plays an integral role in our work, we go into some detail concerning the system architecture. We refer the reader to the paper introducing $\delta$ILP~\cite{dilp1} for more details. 

The $\delta$ILP system provides a framework for differentiable \textit{learning from entailment} ILP. Logic programs are represented by vectors whose components encode whether a particular code fragment is likely to be part of the solution. Such programs are ``executed'' in fuzzy logic, using a weighted average of fuzzy evaluations of code fragments. The hypothesis space consists of all possible combinations of these code fragments captured by user-provided \textit{templates}, a generalization of \textit{metarules}~\cite{metagol}. In our setting \textit{metarules} take the following form:
\begin{definition}[$V$-Metarule]
Let $x,y,z_1,z_2,z_3,z_4$ be first-order variables, $V$ a set of first-order variables such that $x,y\in V$ and $z_1,z_2,z_3,z_4\not \in V$, and $P, Q,$ and $R$ second-order variables ranging over predicate symbols. Then $P(x,y)\mbox{:-} Q(z_1,z_2), R(z_3,z_4)$ is a $V$-\textit{metarule}.
\end{definition}
\noindent Unlike standard metarules,  $V$-Metarules can have various instances. 

\begin{definition}[$V$-Metarule Instance]
Let $M$ be a $V$-Metarule of the form $P(x,y)\mbox{:-} Q_1( $ $x_1,x_2),$ $R_1(x_3,x_4)$. Then 
$P(x,y)\mbox{:-}$ $ Q_1( $ $y_1,y_2),$ $R_1(y_3,y_4)$
is an \textit{Instance} of $M$ if $y_1,y_2,y_3, $ $y_4\in V$.
\end{definition}

Essentially, $V$-metarules are a generalization of metarules that allows for different variable configurations. A $(V,p)$-template is defined using $V$-Metarules as follows:

\begin{definition}[$(V,p)$-template]
Let $p$ be a predicate symbol, $M=P(x,y)\mbox{:-}$ $ Q_1(x_1,x_2),$ $R_1(x_3,x_4)$ and  $M'= P'(x,y)\mbox{:-}$ $ Q_2(x_1,x_2),$ $ R_2(x_3,x_4)$ be $V$-metarules. Then the $(V,p)$-template constructed from $M$ and $M'$ is $(M\{P\mapsto p\},M'\{P'\mapsto p\})$.

An \textit{instance of a $(V,p)$-template} $(M_1,M_2)$ is the pair $(M_1',M_2')$ where $M_1'$ and $M_2'$ are $V$-Metarule Instances of $M_1$ and $M_2$. 

An \textit{instantiation of a $(V,p)$-template $(M_1,M_2)$} is $(M_1'\sigma,$ $M_2'\sigma)$ where $(M_1',M_2')$ is an instance of $(M_1,M_2)$ and $\sigma$ maps the second-order variables to predicate symbols.  
\end{definition}

 Observe that we use templates consisting of two metarules as this is the minimal size, which allows for recursion and disjunction in the predicate definition.

In~\cite{dilp1}, additional restrictions were put on the instantiations of the second-order variables to simplify the hypothesis space for harder tasks. The authors designed the templates to allow precise descriptions of the solution structure, thus simplifying the search. We took a more general approach, which allowed us to define templates uniformly. This results in a larger hypothesis space and, thus, a more challenging experimental setting. 
\begin{example} Consider a $(\{x,y,z\},p)$-template. Possible instances of the contained\\ $(\{x,y,z\})$-metarules include 
$$p(x,y)\mbox{:-} Q_1(x,y), R_1(x,y) \hspace{1.5em} p(x,y)\mbox{:-} Q_2(x,z), R_2(x,z)$$
A program fitting this template would be the following:
\begin{align*}
   p(x,y) :\mbox{-} \mathit{succ}(x,y) \hspace{2em} p(x,y) :\mbox{-} \mathit{succ}(x,z),p(x,z)
\end{align*}
where $R_2$ maps to $p$ and $Q_1$, $R_1$, and $Q_2$ map to $succ$. 
\end{example}
Templates are generalizable to higher-arity predicates; learning such predicates is theoretically challenging for this  ILP setting~\cite{MuggletonLT15}. It is common to restrict learning to dyadic predicates. Reducing template complexity is important when introducing many templates (up to 150). From now on, by template, we mean $(V,p)$-template.

 As input,  $\delta$ILP requires a set of templates $T$ (using pairwise distinct symbols, $p_1,\cdots, p_n$) and \textit{BK}. From the input, it derives a satisfiability problem where each disjunctive clause $C_{i,j}$  denotes the range of possible choices for clause $j$ given template $t\in T$, i.e., overall instantiations of $t$. The logical models satisfying this formula denote logic programs modulo the clauses derivable using the template instantiated by the \textit{BK} and the symbols $p_1,\cdots, p_n$. Switching from a discrete semantics over $\{ 0,1\}$ to a continuous semantics allows the use of differentiable logical operators when implementing differentiable deduction. Solving ILP tasks, in this setting, is reduced to minimizing loss through gradient descent. 

$\delta$ILP uses  $E^+$ and $E^-$ as training data for a binary classifier to learn a model attributing \textit{true} or \textit{false} to ground instances of predicates. This model implements the conditional probability $p(\lambda\vert \alpha, W, T, L,\mathit{BK})$, where $\lambda\in\{\textit{true},\textit{false}\}$, $\alpha$ is a ground instance, $W$ a set of weights, $T$ the templates, and $L$ the symbolic language used to describe the problem containing a finite set of atoms.

Each $(\{x,y,z\},p_i)$-template  $(t_1,t_2)\in T$ is associated with a weight matrix whose shape is $d_1\times d_2$ where $d_j$ denotes the number of clauses constructible using the $\mathit{BK}$ and $L$  modulo the constraints of $t_j$. The number of weights may be roughly approximated  (\textit{quintic}) in terms of the number of templates (considering possible instances and the four second-order variables to instantiate). The weights denote $\delta$ILP's confidence in an instantiation of a template being part of the solution; so-called \textit{per template} assignment. We provide a detailed discussion of weight assignment in Section~\ref{sec:contrib}.

$\delta$ILP implements differentiable inferencing by providing each clause $c$ with a function $f_{c}: [0,1]^m\rightarrow [0,1]^m$ whose domain and range are valuations of grounded instantiations of templates. Note, $m$ is not the number of templates; rather, it is the number of groundings of each template, a much larger number dependent on the \textit{BK}, language bias, and the \textit{atoms} of the symbolic language $L$. Consider a template $(t_1,t_2)$ admitting the clause pair $(c_1,c_2)$, and let the current valuation be $\mathcal{EV}_i$ and $g:[0,1]\times [0,1]\rightarrow [0,1]$ a function computing $\vee$\textit{-clausal} (disjunction between clauses). Assuming we have a definition of $f_{c}$, then  $g(f_{c_1}(\mathcal{EV}_i),f_{c_2}(\mathcal{EV}_i))$ denotes one step of  \textit{forwards-chaining}. Computing the weighted average over all clausal combinations admitted by $(t_1,t_2)$, using the \textit{softmax} of the weights, and finally performing $\vee$\textit{-step} (disjunction between inference steps) between their sums, in addition to $\mathcal{EV}_i$, results in $\mathcal{EV}_{i+1}$. This process is repeated $n$ times (the number of forward-chaining steps), where $\mathcal{EV}_{0}$ is derived from the $\mathit{BK}$.

The above construction still depends on a precise definition of $f_{c}$. Let $$c_g = p(x,y)\mbox{:-} Q_1(y_1,y_2),Q_2(y_3,y_4)$$ where $y_1,y_2,y_3,y_4\in \{x,y,z\}$. We want to collect all ground predicates $p_g$ for which a substitution $\theta$ into $Q_1,Q_2,y_1,y_2,y_3,y_4$ exists s.t. $p_g \in \{Q_1(y_1,y_2)\theta,Q_2(y_3,y_4)\theta\}$. These ground predicates are then paired with the appropriate grounding of the lefthand side of $c_g$. The result of this process can be reshaped into a tensor emphasizing which pairs of ground predicates derive various instantiations of $p(x_1,x_2)$. In the case of body-only variables, there is one pair per atom in the language. Pairing this tensor with some valuation 
$\mathcal{EV}_{i}$ allows one to compute $\wedge$\textit{-literal} (conjunction between literals of a clause) between predicate pairs. 
As a final step, we compute $\vee$\textit{-exists} (disjunction between variants of literals with body-only variables)  between the variants and thus complete computation of the tensor required for a step of \textit{forward-chaining}. 

Four operations parameterize the above process for conjunction and disjunction. We leave discussion to section ~\ref{sec:method}.
\subsection{Related Approaches}
To the best of our knowledge, three recent investigations are related to $\delta$ILP and build on the architecture. The \textit{Logical Neural Network} (LNN)~\cite{SenCRG22}  uses a similar templating, but only to learn non-recursive \textit{chain rules}, i.e. of the form $p_0(X,Y)\mbox{:-} p_1(X,Z_1),$ $\cdots,p_n(Z_n,Y)$; this is simulatable using $\delta$ILP templates, especially for the short rules presented by the authors. They introduced particular parameterized, differentiable, logical operators that are optimizable for ILP. 

Another system motivated by $\delta$ILP is $\alpha$ilp~\cite{SPD2023}. The authors focused on learning logic programs that recognize visual scenes rather than general ILP tasks. The authors restricted the \textit{BK} to predicates explaining aspects of the visual scenes used for evaluation. To build clauses, the authors start from a set of initial clauses and use a \textit{top-k beam search} to iteratively extend the body of the clauses based on evaluation with respect to $E^+$. In this setting, predicate invention and recursive definitions are not considered. Additionally,  learning a relation between two variables is not considered. 

\textit{Feed-Forward Neural-Symbolic Learner}~\cite{CunningtonLLR23} does not directly build on the $\delta$ILP architecture but provides an alternative approach to one of the problems the $\delta$ILP investigation addressed, namely developing a neural-symbolic architecture that can provide symbolic rules when presented with noisy input. In this work~\cite{CunningtonLLR23}, rather than softening implication and working with fuzzy versions of logical operators, compose the Inductive \textit{answer set programming} learners \textit{ILASP}~\cite{ILASP} and \textit{FASTLAS}~\cite{LawRBB020} with various pre-trained neural architectures for classification tasks. This data is then transformed into a weighted knowledge for the symbolic learner. While this work is to some extent relevant to the investigation outlined in this paper, we focus on improving the differentiable implication mechanism developed by the authors of $\delta$ILP 
rather than completely replacing it. Furthermore, the authors focus on tasks with simpler logical structures, similar to the approach taken by $\alpha$ilp.

Earlier investigations, such as NeuraILP~\cite{YangYC17}, experimented with various \textit{T-norms}, and part of the investigation reported in~\cite{dilp1} studied their influence on learning. The authors leave scaling their approach to larger, possibly recursive programs as future work, a limitation addressed herein. 

In~\cite{ShindoNY21}, the authors built upon $\delta$ILP but further restricted templating to add a simple term language (at most term depth 1). Thus, even under the severe restriction of at most one body literal per clause, they can learn predicates for list \textit{append} and \textit{delete}. Nonetheless, scalability remains an issue. \textit{Neural Logical Machines}~\cite{DongMLWLZ19}, in a limited sense, addressed the scalability issue. The authors modeled propositional formulas using multi-layer perceptrons wired together to form a circuit. This circuit is then trained on many (10000s) instances of a particular ILP task. While the trained model was accurate, interpretability is an issue, as it is unclear how to extract a symbolic expression. Our approach provides logic programs as output, similar to $\delta$ILP. 

Some related systems loosely related to our work are \textit{Logical Tensor Networks}~\cite{DonadelloSG17}, \textit{Lifted
Relational Neural Networks}~\cite{SourekSZSK17}, \textit{Neural theorem prover}~\cite{MinerviniBR0G20}, and \textit{DeepProbLog} ~\cite{ManhaeveDKDR21}. While some, such as Neural theorem prover, can learn rules, it also suffers from scaling issues. Overall,  these systems were not designed to address learning in an ILP setting. Concerning the explainability aspects of systems similar to $\delta$ILP, one notable mention is \textit{Logic Explained Networks}~\cite{CiravegnaBGGLMM23}, which adapts the input format of a neural learner to derive explanations from the output. However, the problem they tackle is only loosely connected to our work.  

\section{Contributions}
\label{sec:contrib}
\begin{figure*}
 \begin{center}
    \begin{subfigure}{0.45\textwidth}
    \begin{lstlisting}[language=Prolog]
e(A,B):-i7(B,B),i7(B,B)
e(A,B):-i7(B,A),i7(A,B)
i7(A,B):-z(C,B),z(C,B)
i7(A,B):-i13(A,C),s(C,B)
i13(A,B):-s(C,B),i7(C,C)
i13(A,B):-z(A,B),z(C,A)
    \end{lstlisting}
    \end{subfigure}
\begin{subfigure}{0.45\textwidth}
\begin{lstlisting}[language=Prolog]
e(A,B):-i2(C,B),i2(C,B)
e(A,B):-i18(B,C),i2(C,A)
i2(A,B):-i2(C,C),i18(B,C)
i2(A,B):-z(A,B),z(A,B)
i18(A,B):-z(C,A),z(A,A)
i18(A,B):-s(B,C),s(C,A)
\end{lstlisting}
    \end{subfigure}

   \begin{subfigure}{0.45\textwidth}
\includegraphics[width=65pt]{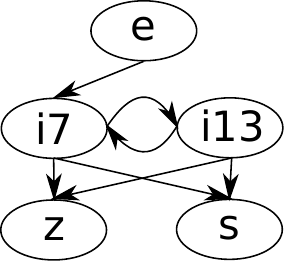}    
    \end{subfigure}  
    \begin{subfigure}{0.45\textwidth}
\includegraphics[width=65pt]{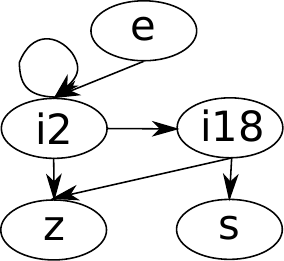}    
    \end{subfigure}
    \end{center}
    \vspace{1mm}
    \caption{\emph{even} ($e$ above) solutions  trimmed to used templates. Note, \textit{s} denotes successor, and $z$ denotes zero.}
    \label{fig:even_samples}
\end{figure*}

 The number of weights used by $\delta$ILP is approximately \textit{quintic} in the number of templates used, thus incurring a significant memory footprint (See Section~\ref{subsec:dilp}). This issue is further exacerbated as computing evaluations requires grounding the hypothesis space. As a result, experiments performed in~\cite{dilp1} use task-specific templates precisely defining the structure of the solution; this is evident in their experiments as certain tasks, i.e., the \textit{even} predicate (See Figure \ref{fig:even_samples}), list multiple results, with different templating. These restrictions result in only a few of the many possible solutions being present in the search space. Furthermore, this \textit{language bias} greatly influences the success rate of $\delta$ILP on tasks such as \textit{length}; the authors report low loss in \textbf{92.5}\% of all runs, which significantly differs from our reduced \textit{language bias} experiments, i.e. \textbf{0}\% correct solutions and \textbf{0}\% achieving low loss.  
 
To achieve low loss, in addition to the chosen \textit{language bias}, $\delta$ILP's authors assign weights \textit{per template} resulting in a large vector $v$ of learnable parameters, see Section~\ref{sec:method}. This seems to imply high dimensionality; however, intuitively, \textit{softmax} is applied to $v$ during differentiable inferencing and thus transforms $v$ into a distribution, effectively reducing its dimensionality. 
 
Our investigation aims to: \textbf{(i)} increase the dimensionality of the search space while maintaining the efficacy of the differentiable inferencing, and \textbf{(ii)} minimize the \textit{bias} required for effective learning. We proceed by adding many (\{x,y,z\},p)-templates (each with a unique symbol $p$) as discussed in subsection~\ref{subsec:dilp}; this largely reduces the biases towards solutions of a particular shape. Nonetheless, given the significant number of weights required, large-scale PI  remains highly intractable. Thus, we amend how weights are assigned to templates (See Section~\ref{sec:method}). 

Assigning weights \textit{per template} is the main source of the significant memory footprint. The authors discuss this design choice in Appendix F of~\cite{dilp1}. In this Appendix, the authors describe assignment \textit{per clause}, i.e., the weight denotes the likelihood of a given instantiation of a metarule instance occurring within an instantiation of a template. This approach was abandoned as it was ``incapable of escaping local minima on harder tasks''. Assigning weights \textit{per clause} results in roughly a quadratic reduction in the number of weights. 
 
Our system ($\delta$ILP$_2$) assigns weights \textit{per literal}, i.e., the weight denotes the likelihood of a given literal occurring within an instantiation of a template. Assignment \textit{per literal} results in another roughly quadratic reduction in the number of assigned weights. 

We observed that the most challenging tasks require learning a binary relation whose solution requires using a third (non-argument) variable, an additional body-only variable. This observation differs from the observations presented in~\cite{dilp1} where complexity was measured purely in terms of the number of \textit{learned predicate definitions} required. For example, consider \textit{0=X mod 5}-hard and \textit{Y=X+4}; our approach solves the former 61\% of the time and the latter 4\% of the time. Note, \textit{0=X mod 5}-hard requires four learned predicate definitions while \textit{Y=X+4} requires two, yet unlike \textit{0=X mod 5}-hard, both predicates relate two variables through a third non-argument variable. 
 
 We test $\delta$ILP$_2$ and  support our observation through experimentally testing the following hypotheses (see Section~\ref{sec:experiments}):

\begin{itemize}
    \item \underline{Hypothesis 1:} Differentiable ILP benefits from increasing the number of templates used during training. 
    \item \underline{Hypothesis 2:} The benefit suggested by \textit{hypothesis 1} is not solely due to the relationship between increasing the number of templates and training a multitude of times with a task-specific number of templates. 
    \item \underline{Hypothesis 3:}  Learning binary predicates using body-only variables remains a challenge regardless of the weight assignment approach. 
\end{itemize}
\section{Methodology}
\label{sec:method}
We now outline the methodological differences between our implementations of differentiable inferencing and $\delta$ILP; We \textbf{(i)}  assign weights \textit{per literal}, \textbf{(ii)}  use slightly different logical operators, \textbf{(iii)} use more precise measures of training outcomes, and \textbf{(iv)}  use a slightly different method of batching examples. 



\subsubsection{Weight assignment} To exploit the benefits of large-scale PI, we need to reduce the large memory footprint incurred by $\delta$ILP's weight assignment, \textit{per template} assignment. We distinguish three types of weight assignment: 
\begin{itemize}
    \item \underline{\textit{per template}}: weight encodes the likelihood that a pair of clauses is the correct choice for the given template.
    \item \underline{\textit{per clause}}: weight encodes the likelihood that a clause occurs in the correct choice of clauses for the given template.
    \item \underline{\textit{per literal}}: weight encodes the likelihood a literal occurs in one of the correct clauses for the given template.
\end{itemize}
 \textit{Per literal} assignment is the coarsest of the three but also has the least memory footprint, thus allowing for large-scale PI. Our system, $\delta$ILP$_2$, implements    \textit{per Literal} assignment.

\subsubsection{T-norm for fuzzy logic} Differentiable inferencing (Section~\ref{sec:relatedwork}) requires four differentiable logic operators. The choice of these operators greatly impacts overall performance. The Author's of $\delta$ILP experimented with various \emph{t-norms}, continuous versions of \textit{classical} conjunction~\cite{ESTEVA2001271}, from which continuous versions of other logical operators are derived. The standard t-norms are \textit{max} ($x\wedge y \equiv \max\{x,y\}$),  \textit{product} ($x\wedge y \equiv x\cdot y$) and Łukasiewicz ($x\wedge y \equiv \max\{x+y-1,0\}$). For simplicity, we refer to all operators derived from a t-norm by \textbf{the conjunctive operator}, i.e., $x\vee y \equiv \min\{x,y\}$ is referred to as \textit{max} when discussing the chosen t-norm.
\begin{center}
\begin{tabular}{r|c|c}
    & $\delta$ILP &  $\delta$ILP$_2$ \\ \hline
    $\wedge$-Literal & product & product \\
    $\vee$-Exists & max & max \\
    $\vee$-Clausal & max & max \\
    $\vee$-Step & product & max
\end{tabular}
\end{center}
When computing many inference steps, \textit{product} produces vanishingly small gradients. Large programs require more inferencing, see Figure~\ref{fig:buzz_small}, thus, we use \emph{max} for $\vee$-step.

\subsubsection{Batch probability}

We require computing values for all predicates over all combinations of atoms, thus motivating an alternative approach to typical \textit{mini-batching}. Instead of parameterization by \textit{batch size}, we use a \textit{batch probability} --  the likelihood of an example contributing to gradient computation. When computing the loss, the example sets $E^+$ and $E^-$ equally contribute. Regardless of the chosen examples, the loss is balanced (divided by the number of examples contributing). If batching results in no examples from $E^+$ ($E^-$), we set that half of the loss to $0$ (with $0$ gradient). Performance degrades when the \textit{batch probability} is near $0.0$ or $1.0$. In our experiments, we used  $0.5$.

\subsection{Considered outcomes}

The experiments outlined in section~\ref{sec:experiments} (See Table~\ref{tab:dilp2}A \& \ref{tab:dilp2}B) allow for five possible outcomes: \textit{Correct on Test (\textbf{C})}, \textit{Fuzzily Correct on Test (\textbf{F})}, \textit{Correct on Training (\textbf{CT})}, \textit{Fuzzily Correct on Training (\textbf{FT})}, and \textbf{FAIL}. We differentiate between test and training to cover the possibility of \textbf{overfitting} and differentiate between correct and fuzzy to cover the possibility of learning programs only correct using fuzzy logic. 

\subsubsection{Overfitting}
$\delta$ILP avoids overfitting as the search space is restricted enough to exclude overfitting programs. However, this is no longer the case when 100s of templates are used. For example, when learning  \textit{even}, it is possible, when training with enough invented predicates, to remember all even numbers provided in $E^+$. Thus, we add a validation step to test our solutions on unseen data (i.e., numbers up to 20 after training on numbers up to 10). Given the types of tasks we evaluated and the structure of the resulting model, a relatively large number of unseen examples is enough to validate. Learning over-fitting solutions for large, unseen input is highly unlikely as the programs would be very large. During experimentation, we observed that $\delta$ILP$_2$ rarely overfits, even when it clearly could. A plausible explanation is that shorter, precise solutions have a higher frequency in the search space.

\subsubsection{Fuzzy solutions}
Another class of solutions observed in both our and earlier experimental designs is \textit{fuzzy solutions}; that is, programs that made correct predictions using fuzzy logic but incorrect predictions when evaluated using classical logic (selected predicates with the highest weight). Typically, fuzzy solutions are worse at generalizing -- they are correct when tested using the training parameters (for example, inference steps) and break on unseen input. Entirely correct solutions for \emph{even} are translatable into a program correct for all numbers, while a fuzzy solution fails to generalize beyond training.

\section{Experiments}
\label{sec:experiments} 
\begin{table}
    \centering
    \renewcommand{\arraystretch}{1.3}
\begin{minipage}{.40\textwidth}
    \scalebox{.8}{
    \begin{tabular}{c|c|c|c|c|c|c}
        \makecell{Task} & 
         \textbf{C} & \textbf{F} & 
         \textbf{CT} & \textbf{FT} &
         \makecell{diff \textbf{C}} & \makecell{diff \textbf{CT}}\\
         \hline
         predecessor\textbf{/2}            &  100      & 100      & 100      & 100  & \textbf{+2\%}$^\dagger$  & \textbf{+2\%}$^\dagger$  \\
         even\textbf{/1}                   &      92       & 99       & 92       & 99   & \textbf{+32\%} & \textbf{+32\%}   \\
         $(X\leq Y)\textbf{/2}^*$          & 30       & 31       & 35       & 38   & -70\% & -65\%   \\
         (0=X mod 3)\textbf{/1}                   &     91       & 97       & 91       & 97   & \textbf{+91\%} & \textbf{+91\%}   \\
         (0=X mod 5)\textbf{/1}-easy     &             77       & 80       & 97       & 100  & \textbf{+77\%} & \textbf{+97\%}    \\ 
         (0=X mod 5)\textbf{/1}-hard      &          61       & 65       & 61       & 65   & \textbf{+61\%} & \textbf{+65\% } \\ 
         (Y=X+2)\textbf{/2}                  &  99       & 100      & 99       & 100  & -1\% $^\dagger$&  -1\% $^\dagger$ \\
         (Y=X+4)\textbf{/2}                &  4        & 12       & 5        & 13   & \textbf{+4\%}$^\dagger$ &  \textbf{+5\%}$^\dagger$ \\
         member\textbf{/2}$^*$             & 17       & 19       & 37       & 43   & -67\% & -67\%  \\
         length\textbf{/2}                 &  25       & 26       & 31       & 38   & \textbf{+25\%} & \textbf{+31\%} \\
        grandparent\textbf{/2}             &  38       & 38       & 92       & 94   & \textbf{+38\%} & \textbf{+89\%}    \\
         undirected\_edge\textbf{/2}        &  94       & 94       & 100      & 100  & \textbf{+75\%} & \textbf{+81\%}    \\
         adjacent\_to\_red\textbf{/1}        &    94       & 99       & 94       & 99   & \textbf{+48\%} & \textbf{+48\%}   \\
         two\_children\textbf{/1}           &     74       & 100      & 74       & 100  & \textbf{+13\%} & \textbf{+13\%}   \\
         graph\_colouring\textbf{/1}        &      83       & 85       & 96       & 100  & -15\% & -2\% $^\dagger$   \\
         connectedness\textbf{/2}$^*$         & 40       & 41       & 98       & 99   & \textbf{+16\%} & \textbf{+74\%}  \\
         cyclic\textbf{/1}                 &            19       & 19       & 90       & 100  & \textbf{+18\%} & \textbf{+89\%}  \\
         \multicolumn{7}{c}{\textbf{(A)}}
    \end{tabular}}
        \end{minipage}
    \begin{minipage}{.08\textwidth}
    \
    \end{minipage}
\begin{minipage}{.40\textwidth}
    \scalebox{.8}{
    \begin{tabular}{c|c|c|c|c}
        Task & \textbf{C} & \textbf{F} & 
         \textbf{CT} & \textbf{FT} \\
         \hline
         predecessor\textbf{/2}            & 98       & 100      & 98       & 100    \\
         even\textbf{/1}                               & 70       & 94       & 70       & 94     \\
         $(X\leq Y)\textbf{/2}^*$              & 100      & 100      & 100      & 100     \\
         (0=X mod 3)\textbf{/1}                           & 0        & 0        & 0        & 0     \\
         (0=X mod 5)\textbf{/1}-easy              & -        & -        & -        & -    \\
         (0=X mod 5)\textbf{/2}-hard                & -        & -        & -        & -     \\
         (Y=X+2)\textbf{/2}                   & 100      & 100      & 100      & 100     \\
         (Y=X+4)\textbf{/2}                & 0        & 0        & 0        & 0     \\
         member\textbf{/2}$^*$                 & 84       & 100      & 84       & 100     \\
         length\textbf{/2}                  & 0        & 0        & 0        & 0     \\
        grandparent\textbf{/2}             & 0        & 0        & 3        & 3     \\
         undirected\_edge\textbf{/2}         & 19       & 100      & 19       & 100    \\
         adjacent\_to\_red\textbf{/1}             & 46       & 90       & 46       & 90     \\
         two\_children\textbf{/1}                & 61       & 100      & 61       & 100   \\
         graph\_colouring\textbf{/1}              & 98       & 100      & 98       & 100   \\
         connectedness\textbf{/2}$^*$         & 24       & 100      & 24       & 100   \\
         cyclic\textbf{/1}                    & 0        & 0        & 1        & 1  \\
         \multicolumn{5}{c}{\textbf{(B)}}
    \end{tabular}}
    \end{minipage}
    \caption{\textbf{(A)} \textit{Per Literal} results: difference computed with respect to Table~\ref{tab:dilp2} (B). Significance computed using t-test and $p< e^{-4}$. We use $\dagger$ to denote differences that are not significant. Tasks are denoted \textit{name}\textbf{/}\textit{arity}, i.e. task \textit{even} has arity 1 and \textit{length} has arity 2. \textbf{(B)} \textit{Per Template} result. Due to significant memory requirements, neither \textit{(0=X mod 5)} task fits in GPU memory (~46GB). Problems annotated with * are easy for \textit{Per template} as the hypothesis space fits in one weight matrix.
    }
    \label{tab:dilp2}
\end{table}

We compare $\delta$ILP$_2$ (\textit{per Literal}) to $\delta$ILP (\textit{per Template}) on tasks presented in~\cite{dilp1}\footnote{Section 5 and Appendix G of~\cite{dilp1}.} plus additional tasks to experimentally test \textit{Hypothesis 2 \& 3}. The tasks are separated into four domains: \textit{numeric}, \textit{list}, \textit{ancestors}, and \textit{graphs}. Results are shown in Table \ref{tab:dilp2}A~\& \ref{tab:dilp2}B. Tasks annotated by \textit{easy} contain extra background knowledge, simplifying the learning process, while \textit{hard} versions do not use the extra background knowledge. Concerning \textit{experimental parameters}, we ran $\delta$ILP$_2$ (\emph{Per Literal} assignment) using 150 templates to produce Table~\ref{tab:dilp2}. For $\delta$ILP  (\emph{Per Template} assignment)~\cite{dilp1}, we ran it with the precise number of templates needed to solve the task. Using more templates was infeasible for many tasks due to the large memory footprint of \textit{per template} assignment. In both cases, we used $(\{x,y,z\},p_i)$-templates with pairwise distinct $p_i$. 

Other parameters are as follows: $2k$ gradient descent steps, early finish when loss reaches $10^{-3}$, differentiable inference is performed for $25$ steps,  batch probability of $0.5$, weights are initialized using a normal distribution,  output programs are derived by selecting the highest weighted literals for each template.

\begin{figure}
    \centering
    \includegraphics[width=0.80\textwidth]{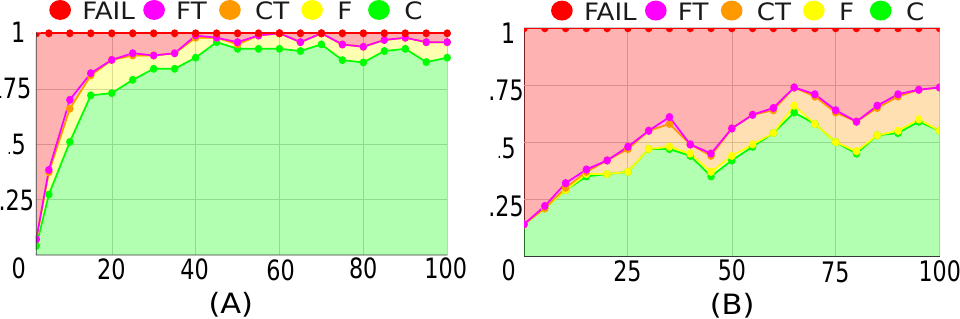}
    \caption{Learning $0=X mod\ 3$ \textbf{(A)} and $X \leq Y$ \textbf{(B)} varying the number of templates (X-axis). The Y-axis is the proportion of solutions in each category. All considered tasks show this pattern.}
    \label{fig:inv}
    \vspace{2em}
\end{figure}

We ran the experiments producing Figure  \ref{fig:inv}  on a computational cluster with 16 nodes, each with 4 GeForce RTX 2070 (with 8~GB of
RAM) GPUs. We ran the experiments producing Table~\ref{tab:dilp2}A~\& \ref{tab:dilp2}B on a GPU server with 8 NVIDIA A40 GPUs (46GB each). We implemented both $\delta$ILP and $\delta$ILP$_2$ using PyTorch~\cite{torch} (version 2.0). Our implementation can be found at the following repository: \href{https://github.com/Ermine516/DILP2}{\underline{{\color{blue}github.com/Ermine516/DILP2}}}.

\subsection{Hypothesis 1}

 Figure \ref{fig:inv} illustrates the proportion of runs within each of our five categories (\textbf{C}, \textbf{CT}, \textbf{F}, \textbf{FT}, \textbf{FAIL}). As the number of templates increases, the proportion of the runs categorized as correct and generalizing increases. This pattern emerges even for tasks that remain hard to learn. Figure \ref{fig:inv} clearly provides strong evidence supporting \textbf{Hypothesis 1}.

When comparing with $\delta$ILP, out of the 17 tasks we tested $\delta$ILP and $\delta$ILP$_2$ on, $\delta$ILP$_2$ showed improved performance on 13 tasks, and the improved performance was statistically significant for 12 of these tasks. Of the four remaining tasks, $\delta$ILP  showed a statistically significant performance difference on two, namely $X\leq Y$ and \textit{member}. Both benefit from  \textit{per template} assignment as the entire search space fits into one weight matrix. Thus, $\delta$ILP is essentially performing brute force search. While one would expect the same issue to occur for \textit{connectedness}, there are fewer solutions to \textit{member} and $X\leq Y$ in the search space than in the case of \textit{connectedness}; it is a more general concept. Thus, even when $\delta$ILP has an advantage, $\delta$ILP$_2$ outperforms it when training on more complex learning tasks.

Notably, $\delta$ILP$_2$ outperforms $\delta$ILP on many challenging tasks. For example, \textit{0=X mod 5} cannot be solved by $\delta$ILP in our experimental setting. In~\cite{dilp1}, low loss was attained only 14\% of the time when $+2$ and $+3$ are in the \textit{BK}~\cite{dilp1}. In contrast, we achieved a 61\% success rate on this task even when the \textit{BK} contained only \textit{successor} and \textit{zero}; for the dependency graphs of a solution learned, see Figure~\ref{fig:buzz_small}.

\subsection{Hypothesis 2}

\begin{figure}

   \begin{subfigure}{0.55\textwidth}
\includegraphics[width=.8\textwidth]{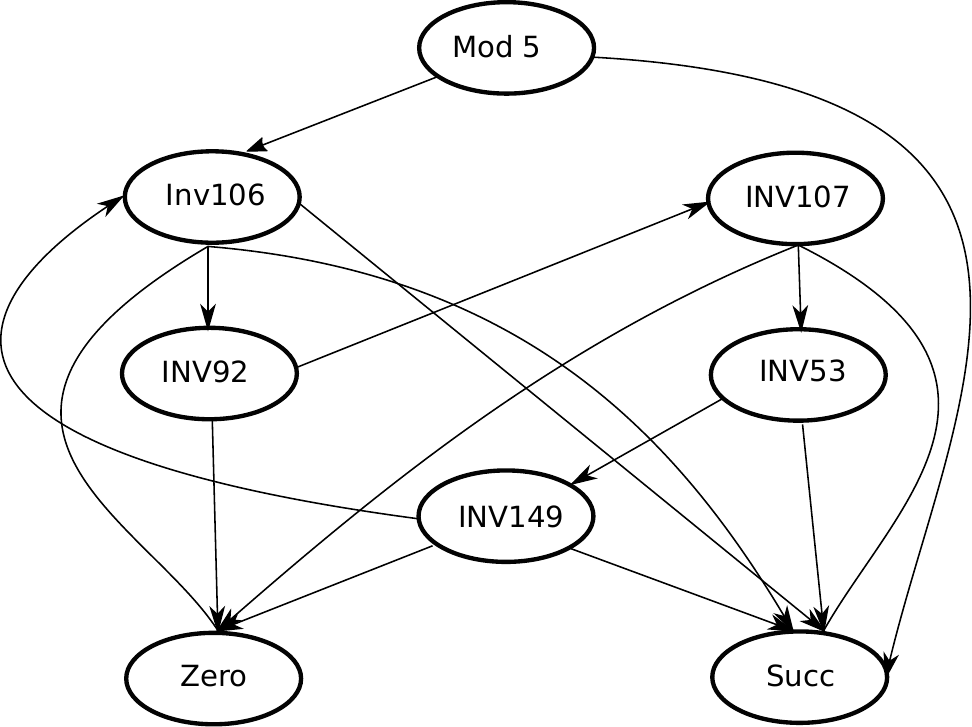}    
    \end{subfigure}  
    \begin{subfigure}{0.4\textwidth}
 \scalebox{.9}{
        \begin{tabular}{c|c|c|c|c}
            task & 
             \textbf{C}& \textbf{F}& 
             \textbf{CT} & \textbf{FT}\\
             \hline
            3 templates & 0\% & 0.02\% & 0\% & 0.02\% \\
            5 templates & 0.11\% & 0.13\% & 0.11\% & 0.14\% \\         
            50 templates & 60\% & 65\% & 60\% & 65\% \\
    \end{tabular}}\vspace{4em}
    \end{subfigure}
    
    \caption{ \textbf{(A)} Template dependency graph of a correct program learned by $\delta$ILP$_2$ for the (0= X mod 5)-hard task. \textbf{(B)} $\delta$ILP$_2$ learning \emph{0 = X mod 6}. Ran $10k$ times for 3 and 5 templates and 100 for 50.}
    \label{fig:buzz_small}
    \vspace{2em}
\end{figure}

To illustrate that large-scale predicate invention is not equivalent to re-initialization of weights, we ran $\delta$ILP$_2$ on the \textit{0=X mod 6}  while varying the numbers of templates used during training (results shown in Figure~\ref{fig:buzz_small}\textbf{B}), i.e. improved performance is not the result of randomly initializing a small number of templates many times. Note, \textit{0=X mod 6} is slightly more challenging than \textit{0=X mod 5} and thus aids in illustrating the effect of larger-scale PI. According to Figure~\ref{fig:buzz_small}\textbf{B}, when training with three templates (minimum required), we would need to run $\delta$ILP$_2$  ~5000 times to achieve a similar probability of success (\textbf{F} solution) as a single 50 predicate run.

When using five invented predicates, which is minimum required to avoid constructing binary predicates (see Hypothesis 3), we would need to run $\delta$ILP$_2$ ~750 times to achieve a similar probability of success (finding a fuzzily correct solution) as a single 50 predicate run.
These results do not necessarily imply that using more predicates is always beneficial over doing multiple runs; however, they show that repeated training with intermediary weight re-initialization is not a sufficient explanation of the observed benefits of large-scale PI.

\subsection{Hypothesis 3}

In Table~\ref{tab:dilp2}\textbf{A} \ \& \ ~\ref{tab:dilp2}\textbf{B}, one can observe that some tasks that require learning a relatively simple program (i.e., \emph{length}) are more challenging than tasks such as \textit{0=X mod 6} that require learning a much larger program.
 
As stated above, we hypothesize this is due to propagating the gradient through an existential quantification. This results in difficulties when learning predicates that relate two input variables through a body-only variable. The difficulty increases with the number of such predicates required to solve the task. 

We introduced a task explicitly designed to test this hypothesis:  ($Y=X+4$). This task requires only one more predicate than $Y=X+2$, yet the success rate drops significantly with respect to $Y=X+2$ (from 99\% to 4\%). For \emph{0=X mod 2} (\emph{even}) and \emph{0=X mod 5}, the change is gradual (from 92\% to 61\%). Thus, the number of relational predicate definitions that a given task requires learning is a more precise measure of complexity than the number of learned predicate definitions. 

\section{Conclusion \& Future Work}
The main contribution of this work (\textit{Hypothesis} 1) is strong evidence that additional templating (beyond what is necessary) improves performance. Verification of this hypothesis used $\delta$ILP$_2$, our modified version of $\delta$ILP. We performed our experiments using reduced language bias compared to the experiments presented in~\cite{dilp1}. Furthermore, we used the same generic template for all predicate definitions learned by the system. This choice makes some tasks significantly more difficult. Additionally, we verified that the performance gains were not simply due to properties shared with weight re-initialization when using a task-specific number of templates during learning (\textit{Hypothesis} 2). 

During experimentation, we noticed that the difficulty of the task did not correlate well with the number of learned predicate definitions needed to solve it but rather with the arity and the necessity of a body-only variable. Therefore, we tested this conjecture using the tasks $Y=X+2$ and $Y=X+4$. While both systems solve $Y=X+2$, performance drastically drops for $Y=X+4$, which only requires learning two invented predicates. Note, \textit{0=X mod 5} requires learning four invented predicates and is easily solved by $\delta$ILP$_2$. This observation highlights the challenging tasks for such synthesis approaches (\textit{Hypothesis} 3) and suggests a direction for future investigation.

As a continuation of our investigation, we plan to integrate ILP with Deep Neural Networks as a hybrid system that is trainable end-to-end through backpropagation. The Authors of $\delta$ILP presented the first steps  in~\cite{dilp1}. One can imagine the development of a network inferring a discrete set of objects in an image \cite{facebook_object_detection}, or integration with Transformer-based \cite{transformer} language models that produce atoms $\delta$ILP$_2$ can process. This research direction can lead to a network that responds to natural language queries based on a datalog database. Also, as part of planned investigations, we consider an ablation study to show that the improved performance is indeed due to large-scale predicate invention and a scalability analysis to test the limits of the approach.





\begin{credits}
\subsubsection{\ackname} Supported by the ERC starting grant no. 714034 SMART, Czech Science Foundation Grant No. 22-06414L, Math$_{LP}$ project (LIT-2019-7-YOU-213) of the  Linz  Institute of Technology and the state of Upper Austria, Cost action CA20111 EuroProofNet, and the Doctoral Stipend of the University of Innsbruck. We would also like to Thank David Coufal (CAS ICS) for setting up and providing access to the institute's GPU Server.

\subsubsection{\discintname}
The authors have no competing interests.
\end{credits}
%
%
%
 \bibliographystyle{splncs04}
%
\bibliography{dilp2}

\end{document}